\documentclass[letterpaper]{article}
\usepackage{aaai23} 
\usepackage{times} 
\usepackage{helvet} 
\usepackage{courier} 
\usepackage[hyphens]{url} 
\usepackage{graphicx} 
\usepackage{graphicx} 
\usepackage{natbib} 
\usepackage{caption} 
\usepackage{mathtools}
\usepackage{amsmath}
\usepackage{subcaption}
\usepackage{wrapfig}
\newcommand{\qizwo}{{\sc QI$^2$}}

\title{ECS - an Interactive Tool for Data Quality Assurance}
\author{
    Christian Sieberichs\textsuperscript{\rm 1},
    Simon Geerkens\textsuperscript{\rm 1},
    Alexander Braun\textsuperscript{\rm{1}},
    Thomas Waschulzik\textsuperscript{\rm 2}}

\affiliations {
    \textsuperscript{\rm 1}University of Applied Sciences Düsseldorf, 40476 Düsseldorf, Germany,\\
    \textsuperscript{\rm 2}Siemens Mobility GmbH, 91058 Erlangen, Germany\\
    \{christian.sieberichs, simon.geerkens, alexander.braun\}@hs-duesseldorf.de, thomas.waschulzik@siemens.com
}
\begin{document}
\maketitle
\begin{abstract}
\begin{quote}
With the increasing capabilities of machine learning systems and there potential use in safety-critical systems, ensuring high-quality data is becoming increasingly important. In this paper we present a novel approach for the assurance of data quality. For this purpose, the mathematical basics are first discussed and the approach is presented using multiple examples. This results in the detection of data points with potentially harmful properties for the use in safety-critical systems.
\end{quote}
\end{abstract}

\section{Introduction}
The development of machine learning (ML) based systems has led to a widespread use in research, industry as well as in the everyday life. Even though ML systems show great performance in solving complex tasks, their use is mostly limited to domains, where wrong decisions only have minor consequences. The application of ML systems in high-risk domains currently is problematic due to the needed quality, lack of trustworthiness and the expected legal basis. To give a legal framework for the application of ML systems the European AI act \cite{european_comission_laying_2021} is at the moment under development. Simultaneously, multiple projects from research and industry are dealing with the topic of ML systems in high risk areas, such like "KI-Absicherung" \cite{habli_integrated_2021} and "safetrAIn" \cite{siemens_safetrain_2022}. All of those projects highlight the high requirements that are needed to protect humans from errors made by ML systems. High risk ML systems have to fulfill the requirements according to \cite{european_comission_laying_2021} Chapter 2 "REQUIREMENTS FOR HIGH-RISK AI SYSTEMS" Article 10 "Data and data governance" Point 3: "Training, validation and testing data sets shall be relevant, representative, free of errors and complete". In this paper we introduce a new approach that will contribute to the future fulfillment of this requirement. It is showcased how different relevant aspects of the data can be analysed and how relations between the given data can be used for quality assurance aspects.

The presented approach is part of the QUEEN-method (\textbf{Qu}alitätsgesicherte \textbf{e}ffiziente \textbf{E}ntwicklung vorwärtsgerichteter künstlicher \textbf{N}euronaler Netze, \textit{quality-assured efficient development of neural networks}) \cite{waschulzik_qualitatsgesicherte_1999} which is a comprehensive approach for the development of quality assured neural networks. In the scope of the QUEEN-method two data quality assurance methods were developed, namely \qizwo (\textbf{i}ntegrated \textbf{q}uality \textbf{i}ndicator) \cite{geerkens_anwendung_2021} and ECS (\textbf{e}quivalent \textbf{c}lasses \textbf{s}ets) \cite{sieberichs_anwendung_2021}. These methods were developed simultaneous in close cooperation. In this paper we want to show the mathematical basis and the use of ECS on the topic of quality assurance. The abilities and usage of the \qizwo is covered in another submission \cite{geerkens_qi2_nodate}.

The ECS is particularly used to analyse the local and global composition of data sets. Based on this a wide variate of data quality properties is addressed. Be it the identification of single data points like outliers, false annotations or isolated data or the identification of groups of data points like decision boundaries and local data point groups of identical output. The ECS makes it possible to identify all data points which do not match specifiable conditions. The method itself is thereby created in such a way that interactions between the user and the data are supported in order to simplify and speed up the quality assurance process.  

\section{Related Work/state of the Art}
Despite the fact that data quality and quality assurance are widely necessary and researched, there exists no single general accepted definition. Instead, there are several attempts to define data quality based on current developments. One example is given by \cite{wang_beyond_1996} who define data quality with respect to the intended use of the data. It is argued that data quality has to be a context dependent term to be appropriately used in the context of a given tasks. In addition, the term of "data quality" is split into multiple properties like accuracy, consistency, completeness, safety and more. In \cite{sidi_data_2012} many of these properties are listed and defined separately. In \cite{pipino_data_2002} data quality is additionally split into subjective and objective assessments of data quality.

A general definition on data quality can thereby not be given. Instead is high data quality considered to be data which is fit for its intended purpose \cite{fadahunsi_protocol_2019}. If data quality is used in standards it is typically split into the different properties which have to be analysed separately like in \cite{gualo_data_2021}.

A first step to assure the data quality is the use of descriptive statistics \cite{holcomb_fundamentals_2016}. Herein statistical methods are used to gain greater insights into the given data. Common methods are the visualization via scatter plots and histograms, often combined with the measurement of central tendencies, dispersion and location parameters. Our proposed method extends the descriptive statistical methods, enables the visualization of multiple quality assurance aspects in one plot and enables a direct interaction between the visualization of the quality indicator visualization and the data.

When trying to assure the quality of data, another possible approach is the representation of given data points in lower dimensial space using methods of dimensinality reduction. Commonly used methods are PCA \cite{jolliffe_principal_1990}, tSNE \cite{maaten_visualizing_2008} or UMAP \cite{mcinnes_umap_2020}. These methods often produce representations interpretable by humans if the output dimensionality is chosen to be low enough. However, such methods often result in considerable loss of information. The ECS on the other hand is computed on the original values and takes all the given information into account.

Some approaches try to cover as many dimensions of the data quality as possible. One way to do this is by testing the data against predefined rules and assumptions. An example of such an approach is the pointblank R package \cite{iannone_pointblank_2022} which is created for an agent based data quality assurance. In this package, specific elements of the data are tested against predefined functions. As part of this it can be tested if the data is greater, equal, lower and so on. Another method is given by DEEQU published in \cite{schelter_automating_2018} and \cite{schelter_deequ_2018}. This package allows for assumption based unit tests which can be defined by the user. Tests on specific parameters of the data, similar to those already mentioned with regard to the pointblank package, are possible as well. A last method that should be mentioned here is shown in \cite{heinrich_assessing_2018}. This approach showcases a probability-based method which calculates a value representing the probability that a data set is free of internal errors with respect to entered rules. The entered rules are based on the presence of data of certain values, comparable to the package pointblank.
The main problem in using the mentioned approaches is the large amount of required knowledge about the data to create accurate assumptions. On top of this, the efficient creation of assumptions is only given if the user is aware that the data quality is influenced in some regards. Due to the reliance on the relationship between data points, our methods do not need any assumptions or rules that are to be specified by an user. Instead our approach can be used without any knowledge about the data.

A different approach is the focus on just a single dimension of the data quality. On the topic of outlier detection these are for example density-based algorithms like \cite{breunig_lof_2000}. In this approach, the amount of local neighbouring data points is calculated and the thereby generated local density is compared with the nearest neighbours. Another approach is using the DBSCAN algorithm \cite{ester_density-based_1996} to cluster the given data. Based on this clustering the method proposed by \cite{tran_manh_thang_anomaly_2011} calculates values to identify clusters of minimal sizes. These clusters are then regarded as possible anomalies.
Another data quality property is the detection of possible outliers, which can also be solved by density based clustering. One example of such an algorithm is given by \cite{fawzy_outliers_2013}. This algorithm uses a fixed clustering to identify clusters followed by the computation of cluster distances. The clusters are classified as anomalous based on the inter-cluster distances and the deviation from the mean inter-cluster distance. Two quite similar approaches are \cite{samara_enhanced_2022} and \cite{samara_complete_2023}. Both approaches use a clustering of the given data in a first step. The first one uses the previously mentioned DBSCAN, the second one uses a cluster algorithm named OPTICS \cite{ankerst_optics_1999}. In a second step, anomalous clusters are identified, once based on inverse distance weighting (IDW) and once using the kringing method.
The main advantage of all of the mentioned methods is the reliable calculation of their data quality property. However, due to the methods focus on one specific data quality property, they are only useful if the assumption exists that this property could contain errors. The advantage of our proposed method is that multiple data quality properties may be analyzed with one approach.

\section{Method}
The ECS is based on the idea that a data set can to be split into input data and output data. The input data defines the dimensions of the data, henceforth are called features, which can be used to predict the output features. The amount of all possible inputs creates the input space \textit{I}. Accordingly is the output space \textit{O} created by all possible outputs. To use the ECS properly all feature values have to be numbers. Features which are not created by numbers have to be represented in some way as a number or a combination of numbers.

To start the calculation of the ECS two metrics are needed. These metrics should be chosen in such a way that "similar" data points according to the semantics of the task that has to be solved have a relatively small distance to each other. At the same time "dissimilar" data points should have a relatively large distance. The distances between two data points can be calculated in the input space and in the output space independent from each other. By doing so, it is possible to use different metrics for the distances in \textit{I} and in \textit{O}. Which metric is best suited for the data set depends on the given type of data and the task to be be solved. In the following, the difference between data points in the input space is named input distance $d_{RI}$. Accordingly, the difference between data points in the output space is called output distance $d_{RO}$.

To differentiate between "similar" and "dissimilar" data points, the distances can be separated into different groups. The minimal approach is to create two groups. One group for relatively small distances and another one for relatively large distances. Doing so requires a threshold, which is called $\delta_{in}$ for distances in the input space and $\delta_{out}$ for distances in the output space. These $\delta$ can be absolute distance values or a percentage of the maximum known distance between data points. They are set based on the data quality properties that should be identified and the used data type. By comparing two data points with each other, four possible scenarios can be distinguished:
\begin{itemize}
    \item small input distance - small output distance
    \item small input distance - large output distance
    \item large input distance - small output distance
    \item large input distance - large output distance
\end{itemize}
Each of these scenarios shows a relation between the data points. If for example the distances are both small, than the data points may showcase a common use case with a typical output. A small input distance in combination with a large output distance on the other hand could showcase complex areas in the input space or an outlier. Either way, the identification of data properties based on two data points is not enough. Due to this the following four ECS-sets are calculated. In these sets, the compared data points are saved, which are part of one of the above scenarios.

\begin{equation} \begin{aligned}
ECS\_EE(D) \coloneqq & \{d_c|d_c \in D^2 \wedge d_{RI}(B) \leq \delta_{in} \\ & \wedge d_{RA}(B) \leq \delta_{out}\}
\end{aligned} \end{equation}
\begin{equation} \begin{split}
ECS\_EU(D) \coloneqq & \{d_c|d_c \in D^2 \wedge d_{RI}(B) \leq \delta_{in} \\ & \wedge d_{RA}(B) > \delta_{out}\}
\end{split} \end{equation}
\begin{equation} \begin{split}
ECS\_UE(D) \coloneqq & \{d_c|d_c \in D^2 \wedge d_{RI}(B) > \delta_{in} \\ & \wedge d_{RA}(B) \leq \delta_{out}\}
\end{split} \end{equation}
\begin{equation} \begin{split}
ECS\_UU(D) \coloneqq & \{d_c|d_c \in D^2 \wedge d_{RI}(B) > \delta_{in} \\ & \wedge d_{RA}(B) > \delta_{out}\}
\end{split} \end{equation}

\begin{table}[ht]
    \centering
    \begin{tabular}{c|c}
         $d_{RI}$ &distance in \textit{I}  \\
         $d_{RA}$ &distance in \textit{O} \\
         $\delta_{in}$ &inputdelta \\
         $\delta_{out}$ &outputdelta \\
         $D$ &dataset \\
         $D^2$ &all possible comparisons of data points out of $D$ \\
         $d_1,d_2$ &two data points out of $D$\\
         $d_c$ &combination of two data points $(d_1,d_2)$
    \end{tabular}
    \caption{parameters which are needed to calculate the ECS-sets}
    \label{tab:ECS-set parameter}
\end{table}

Each of the four ECS-sets represents all comparisons between data points which result in one of the four scenarios. Thereby, an \textit{E} showcases a small distance whereas a \textit{U} showcases a large one. The first of the two letters of the ECS-sets represents the input distance and the second represents the output distance. Following this, $ECS\_EU$ contains all data point comparisons which result in a small input and a large output distance. The $ECS\_UE$ on the other hand contains comparisons which result in a large input and a small output distance.

The information of the ECS-sets can be used to analyse the data points for each of the four scenarios. This way, it is possible to identify data points with specific properties. It would, for example, be possible to identify all data points which have the many dissimilar data points in close proximity. It would also be possible to identify data points which showcase small distances in the input and the output space. By doing so, certain areas of the input space can be identified which correlate with a certain outputs of the output space. It could also be possible to identify features which differentiate certain data points from each other.

\begin{figure}
    \centering
    \includegraphics[width=\linewidth]{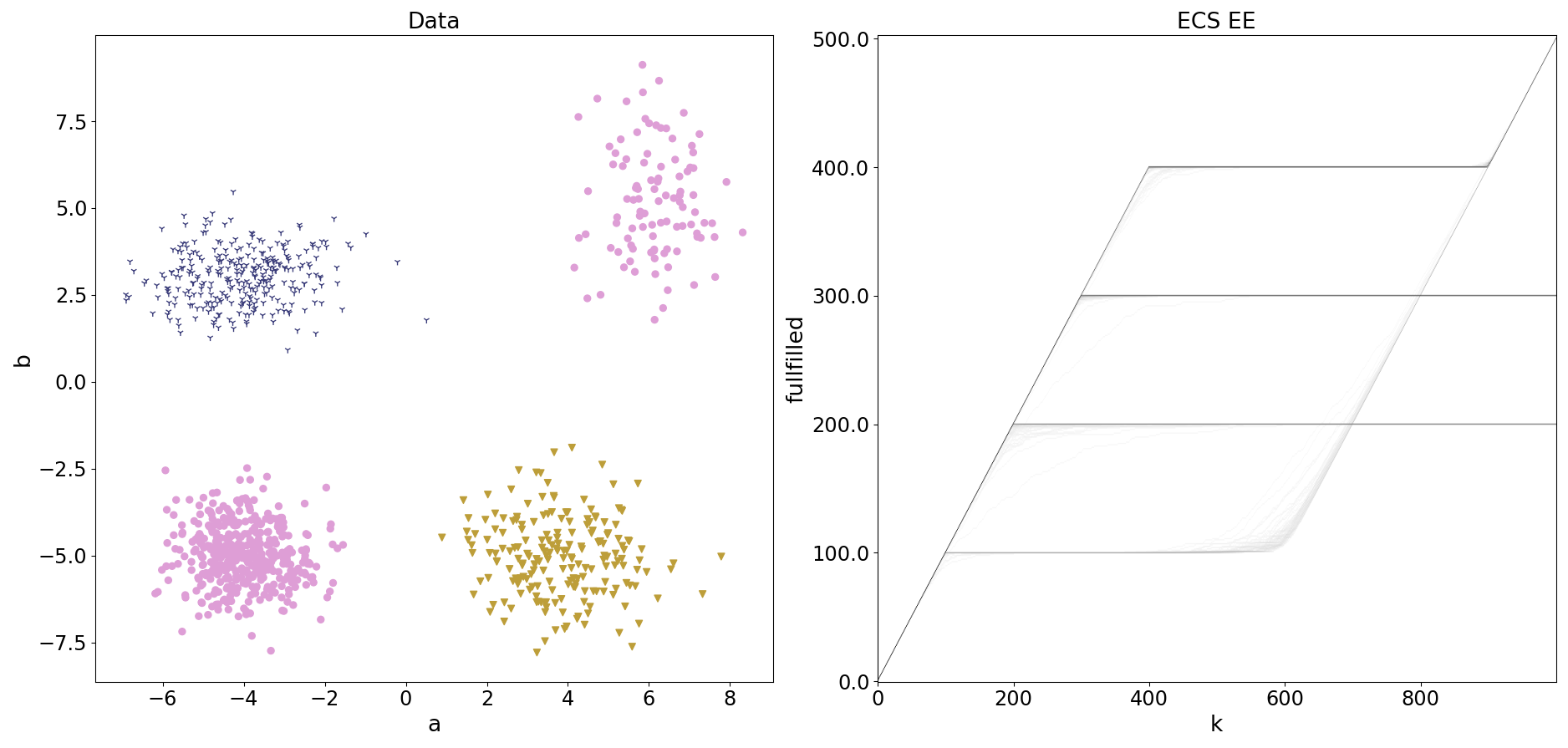}
    \caption{Left: data set created by two input features and one output feature. The data points are located in clusters which can be separated from each other. The cluster in the bottom left and the top right corner are showcasing the same output. Right: $ECS\_EE$ of the displayed data set. Noticeable are steep and early increases which shows the existence of nearby data point with small input and output distance.}
    \label{fig:abb1}
\end{figure}

The ECS-sets contain all information of the data set which could be used for quality assurance. However, the formatting of the sets is difficult for humans to read. This is especially the case when entire data sets should be analysed and not just a small subset of the data. The solution is the comprehensive representation of the ECS-sets in such a way that interesting data points can easily be identified. Before this can be done, it has to be determined which combinations of data points are the most interesting ones. The expectation hereby would be that similar input data would create output data that is related in some way. Based on this, it can be assumed that a combination of data points with a small input distance also has a small output distance. On the other hand it would not be expected that data points with large input distances to each other would showcase similar output data. The most interesting combinations of data points would thereby be combinations which result in a small input distance. This comparisons can be display particularly by sorting the data point comparisons based on the input distance. An example of the sorted representation of the $ECS\_EE$ is shown on the right side of the figure \ref{fig:abb1}. Listed on the x-axis is the comparison between data points. This comparison is data point based and showcases the comparisons of any data point with the $k$th smallest distance in the input space. On the y-axis, it is displayed how many of these comparisons are part of the current ECS-set, which is in this case the $ECS\_EE$. In this process functions are created representing every data point. To showcase the entire data set these functions are superimposed over each other. Every function visually displays if and which of the nearest data points are part of the $ECS\_EE$. The data set which was used to create the displayed $ECS\_EE$ is shown on the left side of figure \ref{fig:abb1}. It is a simple data set created by two input features (a, b) and one output feature (color and shape). Increasing functions display that most of the data points with the $k$th smallest distance are part of the current ECS-set. Functions which do not increase display that the comparisons are part of another ECS-set. It should be emphasized here that the function in the $k$th position only increases for one of the four ECS-sets.

The created ECS-histograms consist of a large number of functions. Areas in the ECS-histogram, in which large amounts of functions showcase the same behavior, are displayed darker. Accordingly smaller amounts of functions are display brighter. For the representation of the amount of functions, gamma correction is used. This way, even singular functions should stay visible.

\begin{figure}
    \centering
    \includegraphics[scale=0.5]{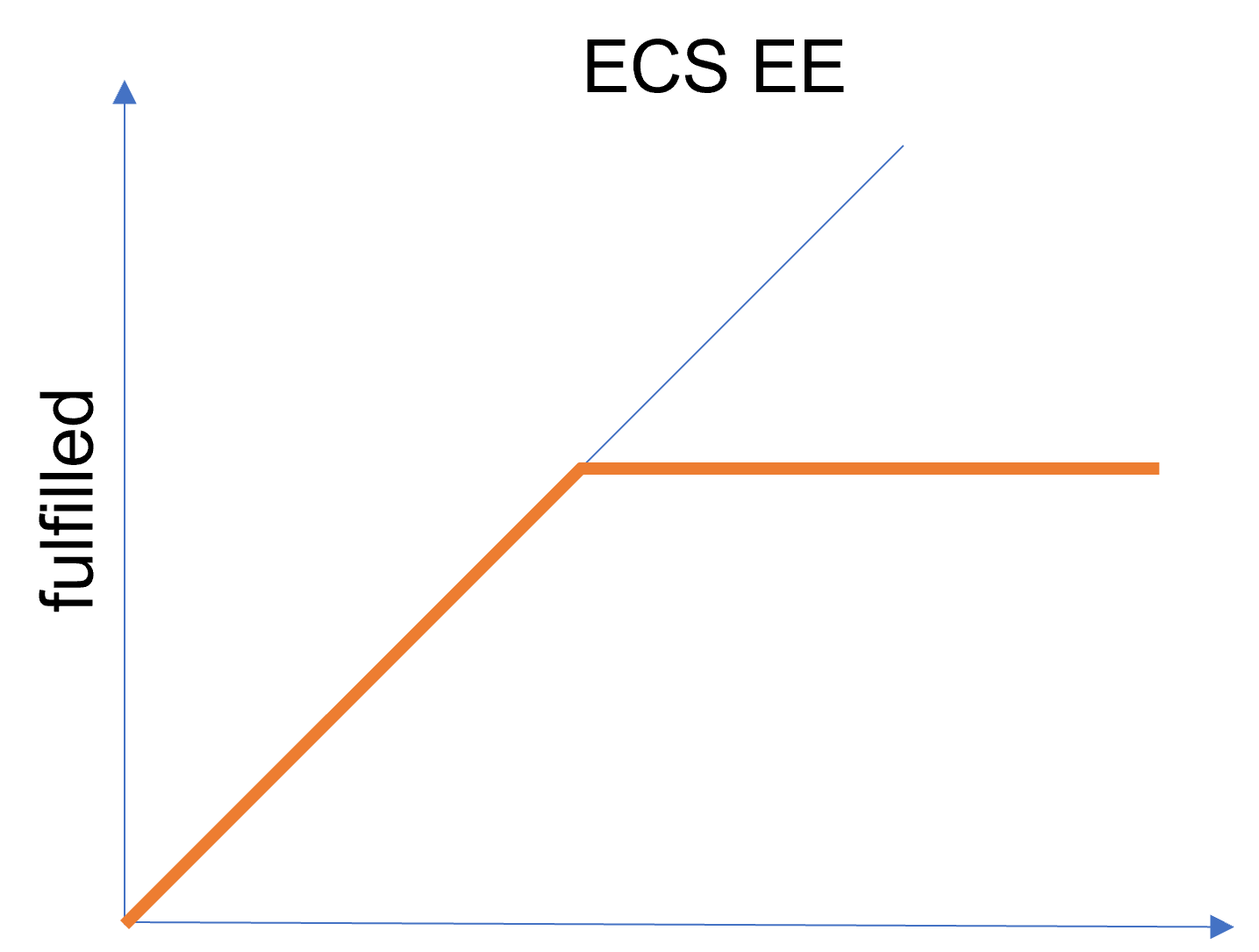}
    \caption{Schematic $ECS\_EE$ created by a data point in an ideal data set.}
    \label{fig:abb2}
\end{figure}

As stated before, it would be expected that a small input distance influences the output distance. An ideal data set would have a strong correlation between the position in the input and the output space. The resulting data point combinations would just have small input and output combinations for all the nearest neighbouring data points. This would result in a steep increase of all functions in the $ECS\_EE$ until all possible similar data points are combined with each other. From this point on the functions in the $ECS\_EE$ do not increase any further. The $ECS\_EE$ function created by a single data point in such an ideal data set is shown schematically in the figure \ref{fig:abb2}. The main diagonal is thereby displaying the maximum speed at which a function is able to increase. Data sets or individual data points which are not ideal do create different functions. One extreme example would be that a function would not increase at all in the $ECS\_EE$. This would be the case because there are no possible combinations with small input distance, small output distance or both. Which of these possibilities is actually the case, can be tested by using the other ECS-histograms.

The benefit of the representation of the information of the data set in the form of the ECS-histogram is the data point based presented information. The neighbouring data points to each data point and there relation to each other are shown and can be compared to set expectations. This basis helps at identifying functions which do not live up to the expectations fairly easy. The reason why the functions are not behaving the way they should, is intrinsically given by the combination of the behaviour and the ECS-set. Additionally, it would be possible to define limits in the ECS-histogram which can be tested autonomously.

\section{Application}
We want to showcase the abilities of the ECS by the application on two different data sets. Simultaneously we want to show how the ECS can be used explicitly to detect certain data quality properties. With this in mind we created a data set which is used as an example. The data set is created in such a way that properties detected by the ECS can be verified by displaying the critical data points. In addition is the ECS used on the MNIST data \cite{lecun-mnisthandwrittendigit-2010} set to display the detection of data quality properties on a commonly known example. The application on the data sets is focused on the the data quality properties created by outliers, isolated data points and local groups of data points with identical output values.

\subsection{ECS on point cloud}
\begin{figure}
    \centering
    \includegraphics[scale=0.35]{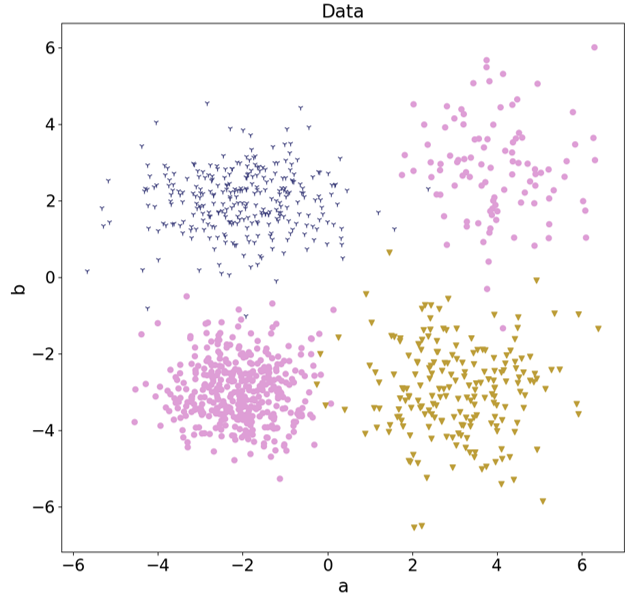}
    \caption{Created data set with two input features, one output feature and 1000 data points. The data points are located in clusters of different density and with different amounts of data points.}
    \label{fig:abb3}
\end{figure}

To demonstrate the usage of the ECS, an artificial data set is created which is displayed at the left side of figure \ref{fig:abb3}. This data set is similar to the one displayed in figure \ref{fig:abb1}. The most important difference is, that the clustered data points can not clearly be separated from each due to the clusters overlapping partially. The data set contains 1000 data points which are grouped in four clusters. As in figure \ref{fig:abb1} each cluster has a different amount and a different density of data points.

The data set was created this way, because it demonstrates a simple classification task. At the same time, properties like outlier and local groups with identical output are present and can be visualized. In the following, it is shown how these properties can be identified by using the ECS. All the ECS-histograms are created using a $\delta_{in}$ of 0.3 times the maximum distance in the input space and a $\delta_{out}$ of 0 to differentiate between all differing outputs.

\subsubsection{outliers}
An outlier is a data point which has an unexpected output for the given input. This output is typically very different from an output that would be expected. Here an outlier is just be considered to vary in the output space. Unwanted variations in the input space are treated in the following section. The reasons for an outlier can be different. The output may for example be wrong or the data point showcases a rare but correct input.

Due to their character, outlier appear in areas which are dominated by data points with a different output. Given this information, it can be stated that an outlier has close neighbours with large output distances. The $ECS\_EU$ is used to identify these cases. Functions in the $ECS\_EU$ are increasing if there are data point combinations with small input and large output distances. Such functions that already increase for the nearest neighbours can thus be regarded as outliers. By targeting these functions, the corresponding outliers can be identified. How many combinations for how many nearest neighbours should be part of the $ECS\_EU$ is dependent on the given data set. In the given point cloud example, 100 nearest neighbours were chosen to be enough to represent the local data points. If out of these 100 combinations more than 70 have a large output distance, then the data point is regarded as an outlier.

\begin{figure}
    \centering
    \includegraphics[width=\linewidth]{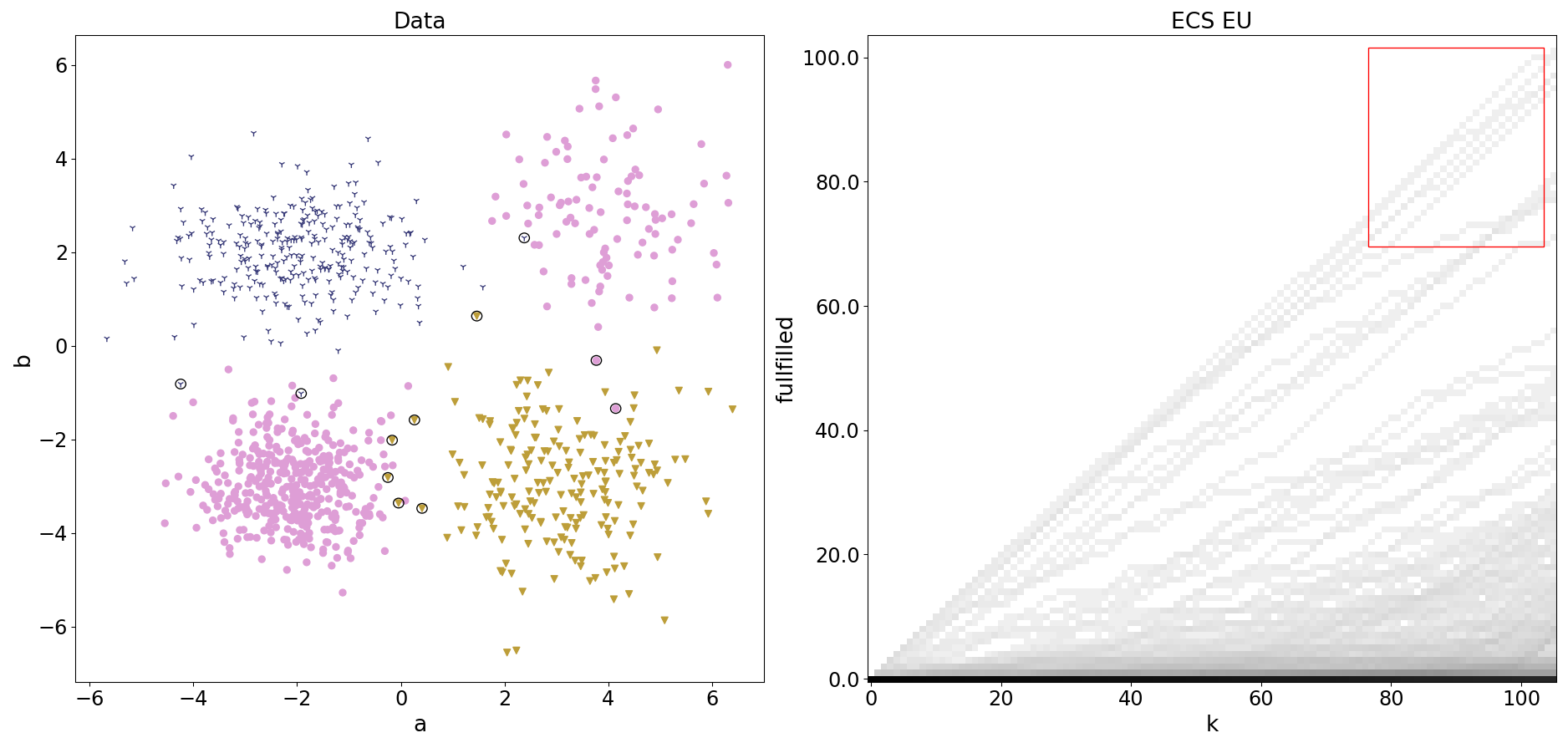}
    \caption{Left: Data set with marked outlier. Right: $ECS\_EU$ of the displayed data set. The area is marked in which functions of outlier appear.}
    \label{fig:abb4}
\end{figure}

The $ECS\_EU$ is shown in figure \ref{fig:abb4}. The area of importance in which the functions of outliers appear is highlighted by a rectangle. It can be noticed that some data points in the point cloud are highlighted which means that they are considered to be outliers. It can also be noticed that most of the functions are not increasing by much.

\subsubsection{isolated data points}
Isolated data points are data points which have a large input distance to many or all of there nearest neighbours. This means that the data point showcases an input which is rare or possibly wrong. In the literature these type of data points are often referred to by "Out-of-distribution-data".

The $ECS\_UE$ and the $ECS\_UU$ are used to identify data points which have large distances to the nearest neighbours. In both ECS-sets are combinations saved which have large input distances to each other. The difference between these two ECS-histograms is the differently sized output distance, which is not considered for isolated data points. The corresponding functions of  isolated data points increase very early in the ECS-histograms. Most of the time, the functions increase in the $ECS\_UE$ as well as in the $ECS\_UU$. This is the case, because the nearest neighbours themselves may have large input distances to each other and thereby showcase very different outputs. The sooner a function increases, the fewer data points are given in the local area of an isolated data point. The amount of neighbouring data points that should exist is on the given task and data set. Typically, this means that every data point should at least have a few neighbouring data points with small input distance. If many data points have no near neighbours, an adjustment of the parameter $delta_{out}$ can be considered.

By using the $ECS\_UE$ and the $ECS\_UU$ it can be stated that, there are no isolated data points in the current data set with less than 50 close neighbours. This can be confirmed by the fact that the sample data set used here was created with clusters of data points



\subsubsection{local groups of identical output}
A local group with identical output is a structure created by multiple data points. All data points in such a group have small distances to each other regarding the input and the output distances. There are no greater amounts of data points which showcase a large output distance, besides possible outliers or false data points. The identification of these groups showcase the ability of the used metric to differentiate between different outputs on the basis of the corresponding input. This means that the input data of the groups share similar features which in turn leads to the differentiation. It would be possible to solve the given task at least for these groups based on these similar features.

The combination of small input distances and small output distances can be identified using the $ECS\_EE$. The functions correlating with data points as part of a local group with identical output increase strongly. The functions will increase as long as there exist data points with small distances in the input and output space in the data set. These strongly increasing functions showcase every data point which is part of such a group of data points. Using the $ECS\_EE$, there is the possibility to also identify groups of different amounts of data points. This can be done by choosing the function increasing the strongest for different amounts of neighbours. If the functions have increase up to the chosen amount, it means that there is a minimum of this amount of data points in the group.

\begin{figure}
    \centering
    \includegraphics[width=\linewidth]{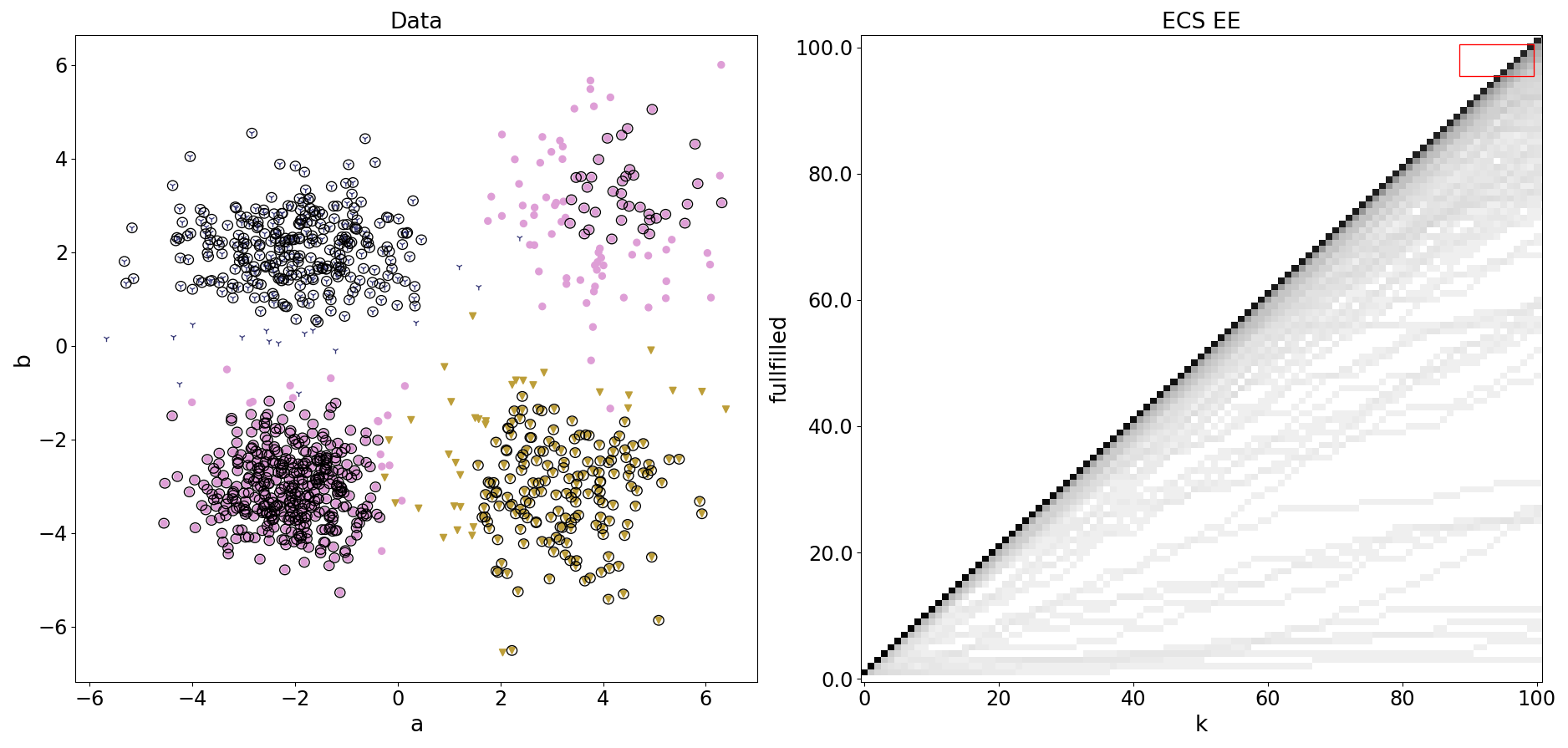}
    \caption{Left: Data set in which local groups of identical output are highlighted. Right: $ECS\_EE$ of the data set with a marked area in the top right for the functions which represent data points in local groups with identical output.}
    \label{fig:abb6}
\end{figure}

In the given case in figure \ref{fig:abb6}, groups with 100 data points and identical output should be identified. The area of importance in the $ECS\_EE$ is marked by an rectangle in the upper right corner. It should be noticed that not just the function increasing the strongest were marked but also some functions which increase a little bit slower. This has be done to make the identified groups more robust against false data points and outliers. In the given case, this means that also functions with 95 out of 100 data points are regarded as local groups with identical output. In addition, it can be noticed that most of the functions in figure \ref{fig:abb6} are increasing. This indicates that there are many data points arranged in local groups. This is logical because the point cloud was created this way as four groups of clustered data points. The detected data points which are part of a local group are marked on the left side of the figure \ref{fig:abb6}.

\subsection{ECS on MNIST}
In contrast to the previous example, MNIST is a data set which was created to represent the specific task of classifying handwritten numbers. The data set consists of 60000 images of size $28*28$ serving as the input and as many numbers between zero and nine for classifying the input.

The most important difference between the previous used point cloud and the MNIST data set is the amount of data and input features. The much greater amount leads to much more functions in the ECS-histograms. The ECS-histograms thereby get more complicated. This can be counteracted by applying more specific metrics to the data. Here the pixel-wise euclidean distance is chosen as a metric. The euclidean distance is typically not used on images due to its bad performance. But in the case of MNIST, this metric is applicable, as the image pixels are given as centered grayscale values. It can be shown that the abilities of the ECS are still given using the euclidean metric. Another problem which appears by using data with many features is the curse of dimensionality through which all distances are getting closer to each other. As a result a larger $\delta_{in}$ of 0.75 times the maximum distance in the input space is used in the following. The $\delta_{out}$ is still 0 to differentiate between all differing outputs.

To show the input data of the MNIST data set in a way, a representation is used in the following chapters. This representation is created by using UMAP \cite{mcinnes_umap_2020}, a dimensionality reduction method. The cluster, which are created this way are marked with a number to showcase the corresponding output. The ECS is used based on the original MNIST input and output data.

\subsubsection{outliers}
\begin{figure}
    \centering
    \includegraphics[width=\linewidth]{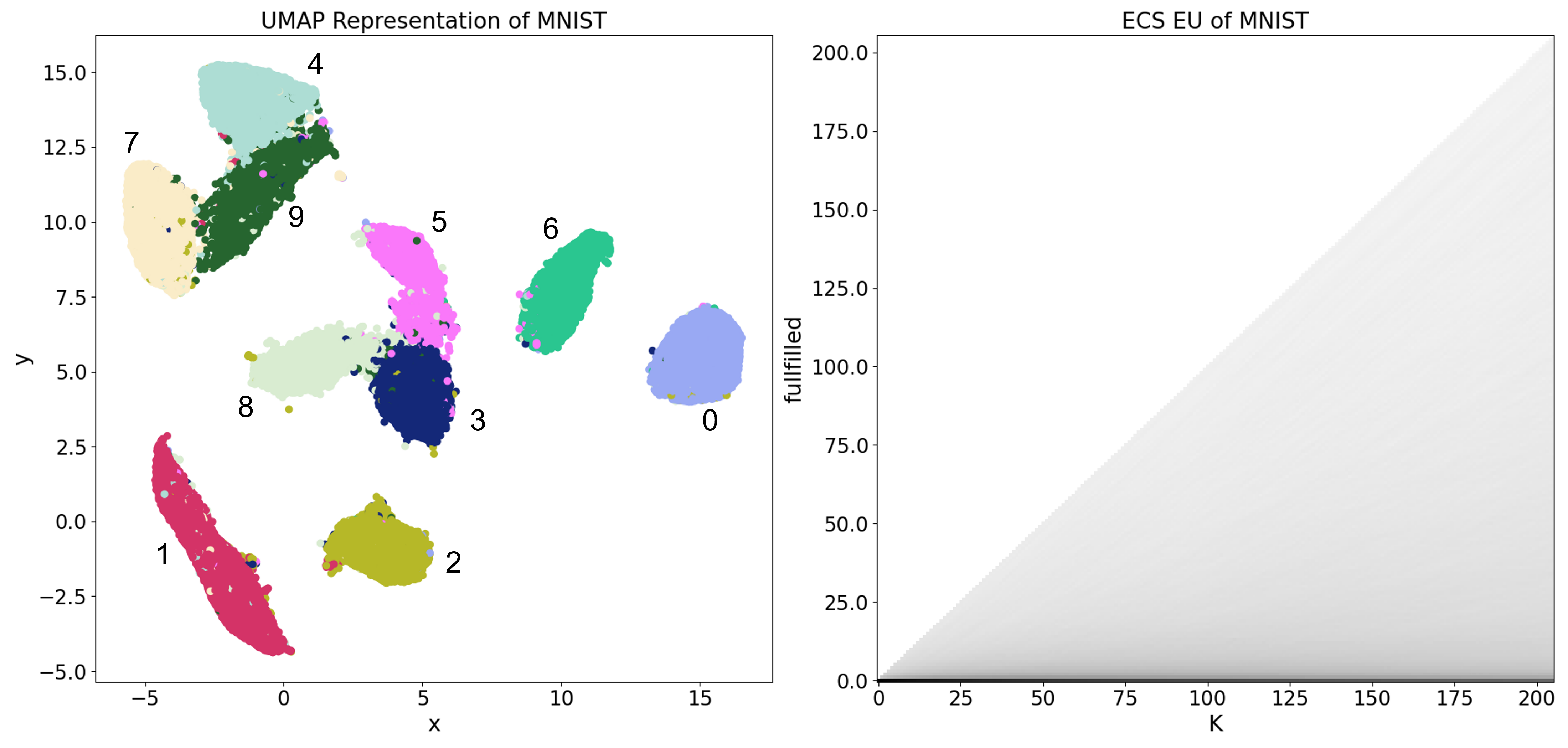}
    \caption{Left: Representation of the MNIST data set with numbers which showcase the output value of the clusters. Right: $ECS\_EU$ which is used to identify outlier.}
    \label{fig:abb7}
\end{figure}

As shown in the chaper "ECS on point cloud - outliers", the $ECS\_EU$ used to identify outliers. The $ECS\_EU$ of the MNIST data set for the nearest 200 neighbours is shown in figure \ref{fig:abb7}. As mentioned before, this representation has much more functions. These are too dense for any singular function to be identified without an interaction. But it can be noticed that most functions do not increase by a lot, as indicated by the darker visualization. The amount of functions ($|F|$) which increase to a specific value of fulfillment ($v_f$) until the 200th neighbour are shown in the following table \ref{tab:MNIST outlier}.

\begin{wraptable}{r}{0.35\linewidth}
    \centering
    \begin{tabular}{c|c}
         $v_f$&$|F|$  \\
         \hline \\
         101-200 &6021 \\
         51-100 &7337 \\
         11-50 &14914 \\
         0-10 &31728 \\
         0 &16813
    \end{tabular}
    \caption{Amount of data point combinations which are part of the $ECS\_UE$ for the 200 nearest neighbours.}
    \label{tab:MNIST outlier}
\end{wraptable}

\begin{figure}
    \centering
    \includegraphics[width=\linewidth]{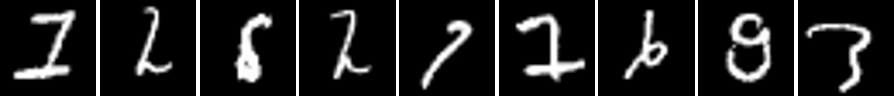}
    \caption{Detected outlier in the MNIST data set. The output of these data points is from left to right: 1,2,5,2,7,7,6,9,3}
    \label{fig:abb8}
\end{figure}

It must be noticed that more than half of the given data points have a maximum of 10 combinations showcasing a small input distance and a large output distance. On the other hand, there are more than 6000 data points for which half of the nearest 200 neighbours have a large output distance. Not all of these are outliers, some may be positioned between classes others may have badly assigned distances. To identify outliers only functions performing worse than a random assignment of distances are used. This means that all data points having more than 180 data points with large output distance among the nearest 200 neighbours are interpreted as outliers. By choosing these functions 804 data points where identified as outliers. In figure \ref{fig:abb8}, a random sample of nine of these outliers is displayed. It is noticeable that all of these data points do look strange. Most may also be mistaken with a different number. It would for example be possible to remove these data points from the MNIST data set to achieve a higher data quality.

\subsubsection{isolated data points}
\begin{figure}
    \centering
    \includegraphics[width=\linewidth]{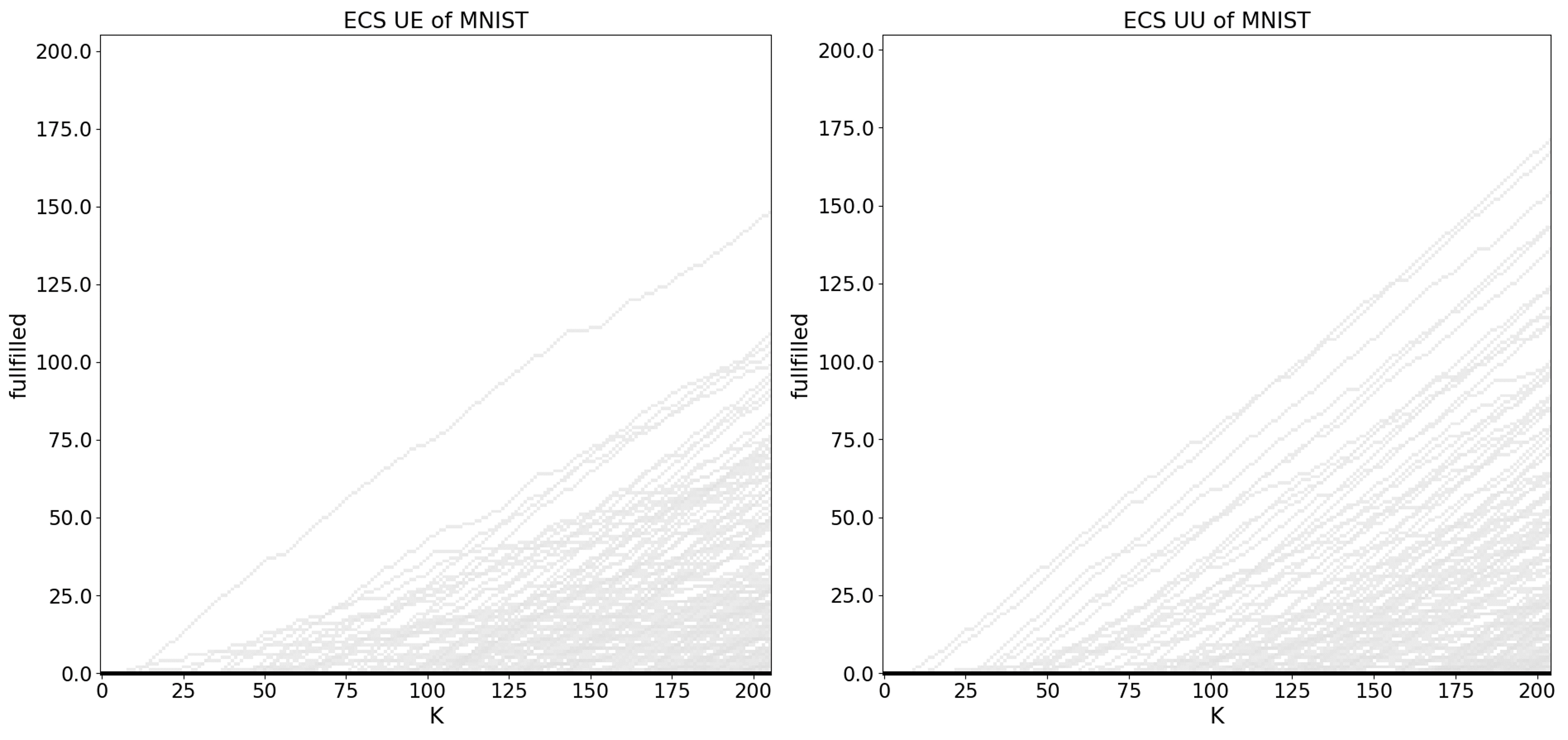}
    \caption{$ECS\_UE$ and $ECS\_UU$ of the MNIST data set for the detection of isolated data points.}
    \label{fig:abb9}
\end{figure}

The identification of isolated data points in MNIST is identical with the identification of data points in the point cloud. The $ECS\_UE$ and $ECS\_UU$ used are shown in figure \ref{fig:abb9}. In the $ECS\_UE$ are 129 and in the $ECS\_UU$ 132 data points with less then 200 neighbouring data points. Most of the correlated data points appear in the $ECS\_UE$ as well as in the $ECS\_UU$. Due to the relatively small amount of increasing functions the histogram is created darker.

\begin{figure}
    \centering
    \includegraphics[width=\linewidth]{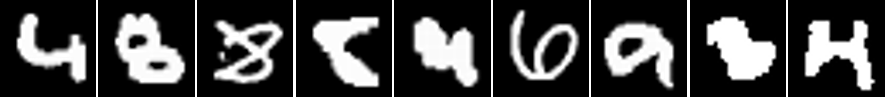}
    \caption{Detected isolated data points in the MNIST data set.}
    \label{fig:abb10}
\end{figure}

The earliest functions start increasing in the $ECS\_UE$ and $ECS\_UU$ for less than 10 neighbours. The input data of the earliest increasing functions is shown in figure \ref{fig:abb10}. One function in the $ECS\_UE$ is noticeable do to its steep and early increase. The corresponding input data is shown in figure \ref{fig:abb10} in the third image from left. This data point has a large distance to its closest neighbours. At the same time, most of these neighbours have the same output "4". This lead to the conclusion that the data point has still some of the most important features which are correlated with the output "4", even if the data point is very isolated. Overall it is noticeable that most isolated data points shown use many input pixels to display the number. This is not often the case in the MNIST data set. In addition, the pixel-wise euclidean distance used reacts especially on pixel-wise differences by assigning higher distances in the input space.

\subsubsection{local groups of identical output}
\begin{figure}
    \centering
    \includegraphics[width=\linewidth]{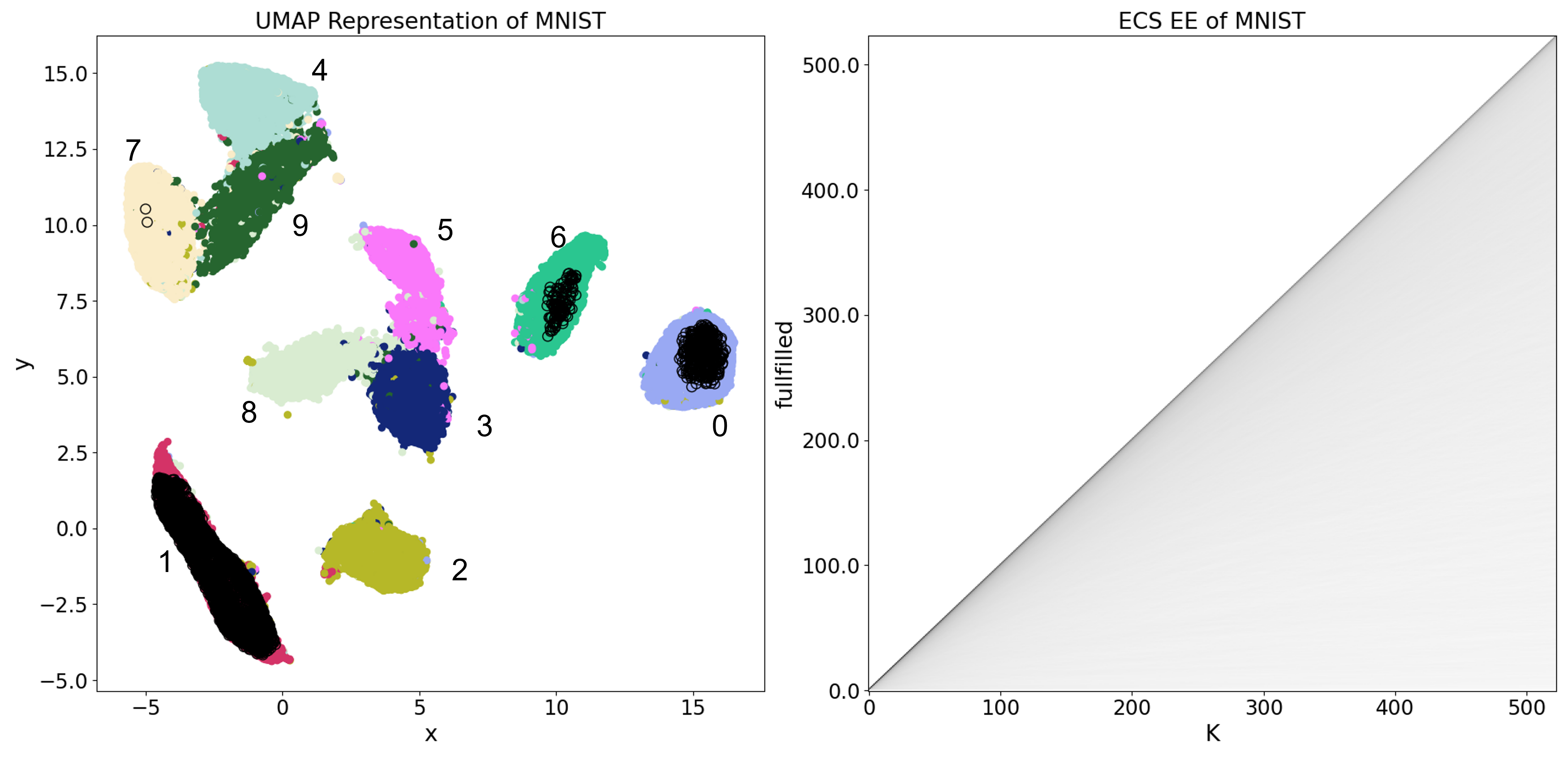}
    \caption{Left: Representation of the MNIST data set with number which showcase the output value of the clusters. Data points which are part of a local group with 1500 data points or more are highlighted in dark. Right: $ECS\_EE$ for the MNIST data set for the nearest 500 neighbours.}
    \label{fig:abb11}
\end{figure}

The $ECS\_EE$, which is used for the identification of local groups of data points with identical output, is shown on the right side of figure \ref{fig:abb11} for the nearest 500 neighbours. As it is the case for the detection of outlier, the amount of functions is much larger than in the point cloud example. It is also not possible to identify single functions but instead overall trends of the functions. It can be noticed that most of the functions are increasing very steeply. This means that most data points do have small input as well as output distances in combination with their nearest neighbours. This in turn means that most data points are located in local groups with identical output. The amount of data points which should be part of the groups can be changed by using different amounts of neighbours in the $ECS\_EE$. In table \ref{tab:MNIST loacal groups}, the amount of data points ($|dp|$) which belong to a local group of different size ($gs$) is shown. These amounts where created by allowing a maximum of 5 data points a different output which could exist due to outliers.

Noticeable are most of the data points located in groups of a few hundred data points. But still more than 4000 data points could be detected which are part of local groups with more than 1500 data points.
\begin{wraptable}{r}{0.45\linewidth}
    \centering
    \begin{tabular}{c|c}
         $gs$ &$|dp|$  \\
         \hline \\
         100 &38383 \\
         200 &27351 \\
         500 &14745 \\
         1000 &7851 \\
         1500 &4329
    \end{tabular}
    \caption{Amount of data points which are part of the different sized local groups with identical output. The entire data set contains 60000 data points.}
    \label{tab:MNIST loacal groups}
\end{wraptable}
The position of these data points is highlighted in dark in the UMAP representation in figure \ref{fig:abb11}. It can be noticed, that especially data points with an output of 1, but also of 0 and 6 show local groups of identical output. This means, that the used metric has the ability to differentiate these data points from each other. The local groups which can be identified can than be used to solve the given task, based on there location in the input space.

\section{Conclusion}
In this paper we presented a novel approach for the data quality assurance based on local similarities. It was shown how the ECS is calculated and can be used on an artificial example. The thereby presented procedure was used to detect data quality properties on the MNIST data set. Besides the possibility to detect outlier, isolated data points and local groups of similar output has the versatile applicability of the ECS been shown. ECS could also be used to validate quantitative data set requirements for data quality properties. These can state the minimum amount of elements per group, the amount of outliers or a maximum amount of local groups. Some of these properties, like the amount of accepted outliers in local groups, may be dependent from associated safety requirements and the required safety integrity level.

\bibliography{Sieberichs}

\begin{thebibliography}{28}
\providecommand{\natexlab}[1]{#1}

\bibitem[{Ankerst et~al.(1999)Ankerst, Breunig, Kriegel, and
  Sander}]{ankerst_optics_1999}
Ankerst, M.; Breunig, M.~M.; Kriegel, H.-P.; and Sander, J. 1999.
\newblock {OPTICS}: {Ordering} {Points} to {Identify} the {Clustering}
  {Structure}.
\newblock In \emph{Proceedings of the 1999 {ACM} {SIGMOD} {International}
  {Conference} on {Management} of {Data}}, {SIGMOD} '99, 49--60. New York, NY,
  USA: Association for Computing Machinery.
\newblock ISBN 1-58113-084-8.
\newblock Event-place: Philadelphia, Pennsylvania, USA.

\bibitem[{Breunig et~al.(2000)Breunig, Kriegel, Ng, and
  Sander}]{breunig_lof_2000}
Breunig, M.~M.; Kriegel, H.-P.; Ng, R.~T.; and Sander, J. 2000.
\newblock {LOF}: {Identifying} {Density}-{Based} {Local} {Outliers}.
\newblock In \emph{Proceedings of the 2000 {ACM} {SIGMOD} {International}
  {Conference} on {Management} of {Data}}, {SIGMOD} '00, 93--104. New York, NY,
  USA: Association for Computing Machinery.
\newblock ISBN 1-58113-217-4.
\newblock Event-place: Dallas, Texas, USA.

\bibitem[{Ester et~al.(1996)Ester, Kriegel, Sander, and
  Xu}]{ester_density-based_1996}
Ester, M.; Kriegel, H.-P.; Sander, J.; and Xu, X. 1996.
\newblock A {Density}-{Based} {Algorithm} for {Discovering} {Clusters} in
  {Large} {Spatial} {Databases} with {Noise}.
\newblock In \emph{Proceedings of the {Second} {International} {Conference} on
  {Knowledge} {Discovery} and {Data} {Mining}}, {KDD}'96, 226--231. AAAI Press.
\newblock Event-place: Portland, Oregon.

\bibitem[{{European Comission}(2021)}]{european_comission_laying_2021}
{European Comission}. 2021.
\newblock {LAYING} {DOWN} {HARMONISED} {RULES} {ON} {ARTIFICIAL} {INTELLIGENCE}
  ({ARTIFICIAL} {INTELLIGENCE} {ACT}) {AND} {AMENDING} {CERTAIN} {UNION}
  {LEGISLATIVE} {ACTS}.

\bibitem[{Fadahunsi et~al.(2019)Fadahunsi, Akinlua, O’Connor, Wark,
  Gallagher, Carroll, Majeed, and O’Donoghue}]{fadahunsi_protocol_2019}
Fadahunsi, K.~P.; Akinlua, J.~T.; O’Connor, S.; Wark, P.~A.; Gallagher, J.;
  Carroll, C.; Majeed, A.; and O’Donoghue, J. 2019.
\newblock Protocol for a systematic review and qualitative synthesis of
  information quality frameworks in {eHealth}.
\newblock \emph{BMJ Open}, 9(3): e024722.

\bibitem[{Fawzy, Mokhtar, and Hegazy(2013)}]{fawzy_outliers_2013}
Fawzy, A.; Mokhtar, H. M.~O.; and Hegazy, O. 2013.
\newblock Outliers detection and classification in wireless sensor networks.
\newblock \emph{Egyptian Informatics Journal}, 14(2): 157--164.

\bibitem[{Geerkens(2021)}]{geerkens_anwendung_2021}
Geerkens, S. 2021.
\newblock Anwendung und {Validierung} des {SHLQI}² auf realen {Beispielmengen}
  und neuronale {Netzwerke}.

\bibitem[{Geerkens et~al.(2021)Geerkens, Sieberichs, Braun, and
  Waschulzik}]{geerkens_qi2_nodate}
Geerkens, S.; Sieberichs, C.; Braun, A.; and Waschulzik, T. 2023.
\newblock {QI2} - an {Interactive} {Tool} for {Data} {Quality} {Assurance}.

\bibitem[{Gualo et~al.(2021)Gualo, Rodriguez, Verdugo, Caballero, and
  Piattini}]{gualo_data_2021}
Gualo, F.; Rodriguez, M.; Verdugo, J.; Caballero, I.; and Piattini, M. 2021.
\newblock Data quality certification using {ISO}/{IEC} 25012: {Industrial}
  experiences.
\newblock \emph{Journal of Systems and Software}, 176: 110938.

\bibitem[{Heinrich et~al.(2018)Heinrich, Klier, Schiller, and
  Wagner}]{heinrich_assessing_2018}
Heinrich, B.; Klier, M.; Schiller, A.; and Wagner, G. 2018.
\newblock Assessing data quality – {A} probability-based metric for semantic
  consistency.
\newblock \emph{Decision Support Systems}, 110: 95--106.

\bibitem[{Holcomb(2016)}]{holcomb_fundamentals_2016}
Holcomb, Z. 2016.
\newblock \emph{Fundamentals of {Descriptive} {Statistics}}.
\newblock Routledge, 0 edition.
\newblock ISBN 978-1-351-97033-4.

\bibitem[{Iannone and Vargas(2022)}]{iannone_pointblank_2022}
Iannone, R.; and Vargas, M. 2022.
\newblock pointblank: {Data} {Validation} and {Organization} of {Metadata} for
  {Local} and {Remote} {Tables}.

\bibitem[{Jolliffe(1990)}]{jolliffe_principal_1990}
Jolliffe, I.~T. 1990.
\newblock {PRINCIPAL} {COMPONENT} {ANALYSIS}: {A} {BEGINNER}'{S} {GUIDE} - {I}.
  {Introduction} and application.
\newblock \emph{Weather}, 45(10): 375--382.

\bibitem[{LeCun and Cortes(2010)}]{lecun-mnisthandwrittendigit-2010}
LeCun, Y.; and Cortes, C. 2010.
\newblock {MNIST} handwritten digit database.

\bibitem[{Maaten and Hinton(2008)}]{maaten_visualizing_2008}
Maaten, L. v.~d.; and Hinton, G.~E. 2008.
\newblock Visualizing {Data} using t-{SNE}.
\newblock \emph{Journal of Machine Learning Research}, 9: 2579--2605.

\bibitem[{McInnes, Healy, and Melville(2020)}]{mcinnes_umap_2020}
McInnes, L.; Healy, J.; and Melville, J. 2020.
\newblock {UMAP}: {Uniform} {Manifold} {Approximation} and {Projection} for
  {Dimension} {Reduction}.
\newblock ArXiv:1802.03426 [cs, stat].

\bibitem[{Mock et~al.(2021)Mock, Scholz, Blank, Hüger, Rohatschek, Schwarz,
  and Stauner}]{habli_integrated_2021}
Mock, M.; Scholz, S.; Blank, F.; Hüger, F.; Rohatschek, A.; Schwarz, L.; and
  Stauner, T. 2021.
\newblock An {Integrated} {Approach} to a {Safety} {Argumentation} for
  {AI}-{Based} {Perception} {Functions} in {Automated} {Driving}.
\newblock In Habli, I.; Sujan, M.; Gerasimou, S.; Schoitsch, E.; and Bitsch,
  F., eds., \emph{Computer {Safety}, {Reliability}, and {Security}. {SAFECOMP}
  2021 {Workshops}}, volume 12853, 265--271. Cham: Springer International
  Publishing.
\newblock ISBN 978-3-030-83905-5 978-3-030-83906-2.
\newblock Series Title: Lecture Notes in Computer Science.

\bibitem[{Pipino, Lee, and Wang(2002)}]{pipino_data_2002}
Pipino, L.~L.; Lee, Y.~W.; and Wang, R.~Y. 2002.
\newblock Data {Quality} {Assessment}.
\newblock \emph{Commun. ACM}, 45(4): 211--218.
\newblock Place: New York, NY, USA Publisher: Association for Computing
  Machinery.

\bibitem[{Samara et~al.(2022)Samara, Bennis, Abouaissa, and
  Lorenz}]{samara_enhanced_2022}
Samara, M.~A.; Bennis, I.; Abouaissa, A.; and Lorenz, P. 2022.
\newblock Enhanced efficient outlier detection and classification approach for
  {WSNs}.
\newblock \emph{Simulation Modelling Practice and Theory}, 120: 102618.

\bibitem[{Samara et~al.(2023)Samara, Bennis, Abouaissa, and
  Lorenz}]{samara_complete_2023}
Samara, M.~A.; Bennis, I.; Abouaissa, A.; and Lorenz, P. 2023.
\newblock Complete outlier detection and classification framework for {WSNs}
  based on {OPTICS}.
\newblock \emph{Journal of Network and Computer Applications}, 211: 103563.

\bibitem[{Schelter et~al.(2018{\natexlab{a}})Schelter, Lange, Schmidt, Celikel,
  Biessmann, and Grafberger}]{schelter_automating_2018}
Schelter, S.; Lange, D.; Schmidt, P.; Celikel, M.; Biessmann, F.; and
  Grafberger, A. 2018{\natexlab{a}}.
\newblock Automating large-scale data quality verification.
\newblock \emph{Proceedings of the VLDB Endowment}, 11(12): 1781--1794.

\bibitem[{Schelter et~al.(2018{\natexlab{b}})Schelter, Schmidt, Rukat,
  Kiessling, Taptunov, Biessmann, and Lange}]{schelter_deequ_2018}
Schelter, S.; Schmidt, P.; Rukat, T.; Kiessling, M.; Taptunov, A.; Biessmann,
  F.; and Lange, D. 2018{\natexlab{b}}.
\newblock {DEEQU} - {Data} quality validation for machine learning pipelines.
\newblock In \emph{{NeurIPS} 2018}.

\bibitem[{Sidi et~al.(2012)Sidi, Shariat~Panahy, Affendey, Jabar, Ibrahim, and
  Mustapha}]{sidi_data_2012}
Sidi, F.; Shariat~Panahy, P.~H.; Affendey, L.~S.; Jabar, M.~A.; Ibrahim, H.;
  and Mustapha, A. 2012.
\newblock Data quality: {A} survey of data quality dimensions.
\newblock In \emph{2012 {International} {Conference} on {Information}
  {Retrieval} \& {Knowledge} {Management}}, 300--304. Kuala Lumpur: IEEE.
\newblock ISBN 978-1-4673-1091-8 978-1-4673-1090-1.

\bibitem[{Sieberichs(2021)}]{sieberichs_anwendung_2021}
Sieberichs, C. 2021.
\newblock Anwendung und {Validierung} des {ECS} auf reale {Beispielmengen} und
  neuronale {Netzwerke}.

\bibitem[{{Siemens}(2022)}]{siemens_safetrain_2022}
{Siemens}. 2022.
\newblock safe.{trAIn}: {Forschungsprojekt} für fahrerlosen {Regionalverkehr}.
\newblock \emph{Internationales Verkehrswesen - Das technisch-wissenschaftliche
  Fachmagazin}.

\bibitem[{{Tran Manh Thang} and {Juntae
  Kim}(2011)}]{tran_manh_thang_anomaly_2011}
{Tran Manh Thang}; and {Juntae Kim}. 2011.
\newblock The {Anomaly} {Detection} by {Using} {DBSCAN} {Clustering} with
  {Multiple} {Parameters}.
\newblock In \emph{2011 {International} {Conference} on {Information} {Science}
  and {Applications}}, 1--5. Jeju Island: IEEE.
\newblock ISBN 978-1-4244-9222-0.

\bibitem[{Wang and Strong(1996)}]{wang_beyond_1996}
Wang, R.~Y.; and Strong, D.~M. 1996.
\newblock Beyond {Accuracy}: {What} {Data} {Quality} {Means} to {Data}
  {Consumers}.
\newblock \emph{Journal of Management Information Systems}, 12(4): 5--33.

\bibitem[{Waschulzik(1999)}]{waschulzik_qualitatsgesicherte_1999}
Waschulzik, T. 1999.
\newblock \emph{Qualitätsgesicherte effiziente {Entwicklung}
  vorwärtsgerichteter künstlicher {Neuronaler} {Netze} mit überwachtem
  {Lernen} ({QUEEN})}.
\newblock Ph.D. thesis, Technische Universität München, München.

\end{thebibliography}
\end{document}